\documentclass[letterpaper, 10 pt, conference]{ieeeconf}  

\IEEEoverridecommandlockouts                              

\usepackage{graphics} 
\usepackage{subcaption} 
\usepackage{epsfig} 
\usepackage[linesnumbered,ruled]{algorithm2e}
\usepackage{mathptmx} 
\usepackage{times} 
\usepackage{amsmath} 
\usepackage{amssymb}  
\usepackage{bm}
\usepackage{multirow}
\usepackage[textsize=footnotesize,color=red!60]{todonotes}

\usepackage[hidelinks]{hyperref}

\DeclareMathAlphabet{\mathcal}{OMS}{cmsy}{m}{n}
\DeclareMathAlphabet\mathbfcal{OMS}{cmsy}{b}{n}
\DeclareMathOperator*{\argmax}{argmax}

\newcommand{\kznote}[1]{#1}
\newcommand{\kznotei}[1]{#1}
\newcommand{\kznoteip}[1]{#1}
\newcommand{\kznoteii}[1]{#1}

\newtheorem{definition}{Definition}

\title{\LARGE \bf
From Pixels to Buildings: End-to-end Probabilistic Deep Networks for Large-scale Semantic Mapping}

\author{Author
  \thanks{Institution}
}

\author{Kaiyu Zheng$^{1}$ and Andrzej Pronobis$^{2}$
  \thanks{$^{1}$Kaiyu Zheng is with Computer Science Dept., Brown University, Prov-
idence, RI, USA \texttt{\footnotesize kaiyu zheng@brown.edu}}
  \thanks{$^{2}$Andrzej Pronobis is with Paul G. Allen School of Computer Science
\& Engineering, University of Washington, Seattle, WA, USA as well
as Robotics, Perception and Learning Lab, KTH, Stockholm, Sweden
\texttt{\footnotesize pronobis@cs.washington.edu}. This work was
  supported by Office of Naval Research (ONR) grant no. N00014-13-1-0817 and
  Swedish Research Council (VR) project 2012-4907 SKAEENet.}
\thanks{We would like to express our gratitude to Prof.~Rajesh~P.~N.~Rao for his unwavering support, encouragement, and invaluable advice.}
}

\begin{document}

\maketitle
\thispagestyle{empty}
\pagestyle{empty}

\begin{abstract}
  We introduce TopoNets, end-to-end probabilistic deep networks for modeling
  semantic maps with structure reflecting the topology of large-scale
  environments. TopoNets build a unified deep network spanning multiple levels
  of abstraction and spatial scales, from pixels representing geometry of local
  places to high-level descriptions of semantics of buildings. To this end,
  TopoNets leverage complex spatial relations expressed in terms of arbitrary,
  dynamic graphs. We demonstrate how TopoNets can be used to perform end-to-end
  semantic mapping from partial sensory observations and noisy topological
  relations discovered by a robot exploring large-scale office spaces. Thanks to
  their probabilistic nature and generative properties, TopoNets extend the
  problem of semantic mapping beyond classification. We show that TopoNets
  successfully perform uncertain reasoning about yet unexplored space and detect
  novel and incongruent environment configurations unknown to the robot.  Our
  implementation of TopoNets achieves real-time, tractable and exact inference,
  which makes \kznoteii{these} new deep models a promising, practical
  \kznoteii{solution} to mobile robot spatial understanding at scale.
\end{abstract}


\section{INTRODUCTION}
The ability to make uncertain inferences about \emph{spatial information} is
fundamental for a mobile agent planning and executing actions in large,
unstructured environments~\cite{hanheide2016ai}, such as office buildings,
airports and search and rescue sites. Robots, while exploring their
environments, gather a growing body of knowledge captured at different spatial
locations, scales (from \kznote{places} to buildings), and levels of abstraction (from
sensory data, through place geometry and appearance, up to high-level semantic
descriptions).

While such information is typically incomplete and noisy, it is also
structured according to relationships that govern the human world. Discovering
and leveraging relationships that span local and global spatial scales as well
as multiple levels of abstraction can help improve robustness, resolve
ambiguities, and enable predictions about latent and unobserved
information~\cite{hanheide2016ai}\cite{pronobis2012icra}\cite{aydemir2013tro}.
Unfortunately, such relationships are also complex and noisy, making semantic
mapping a difficult structured prediction problem. Additionally, semantic maps
are dynamic structures, with dependencies often expressed in terms of graphs
containing a different number of nodes and relations for every environment~\kznote{\cite{pronobis2012icra}}.

As a result, most deep approaches to semantic mapping fail to capture and
exploit such relations. In particular, approaches utilizing convolutional neural
networks focus on relationships constrained to local
scenes~\cite{mccormac2017semanticfusion} and require that the number of latent
variables be constant and related through a similar global
structure~\cite{belanger2016icml}. Other approaches compromise on the structure complexity~\cite{mozos2007ras},
introduce prior structural knowledge~\cite{friedman2007ijcai}, or make hard
commitments about values of semantic attributes~\cite{pronobis2012icra}.
Additionally, these methods are \kznote{often assembled from} independent spatial
models~\cite{pronobis2012icra}\cite{brucker2018semantic}, which exchange
information in a limited fashion.

\begin{figure}[t!]
  \centering
  \captionsetup{width=\linewidth}
  \includegraphics[width=\linewidth]{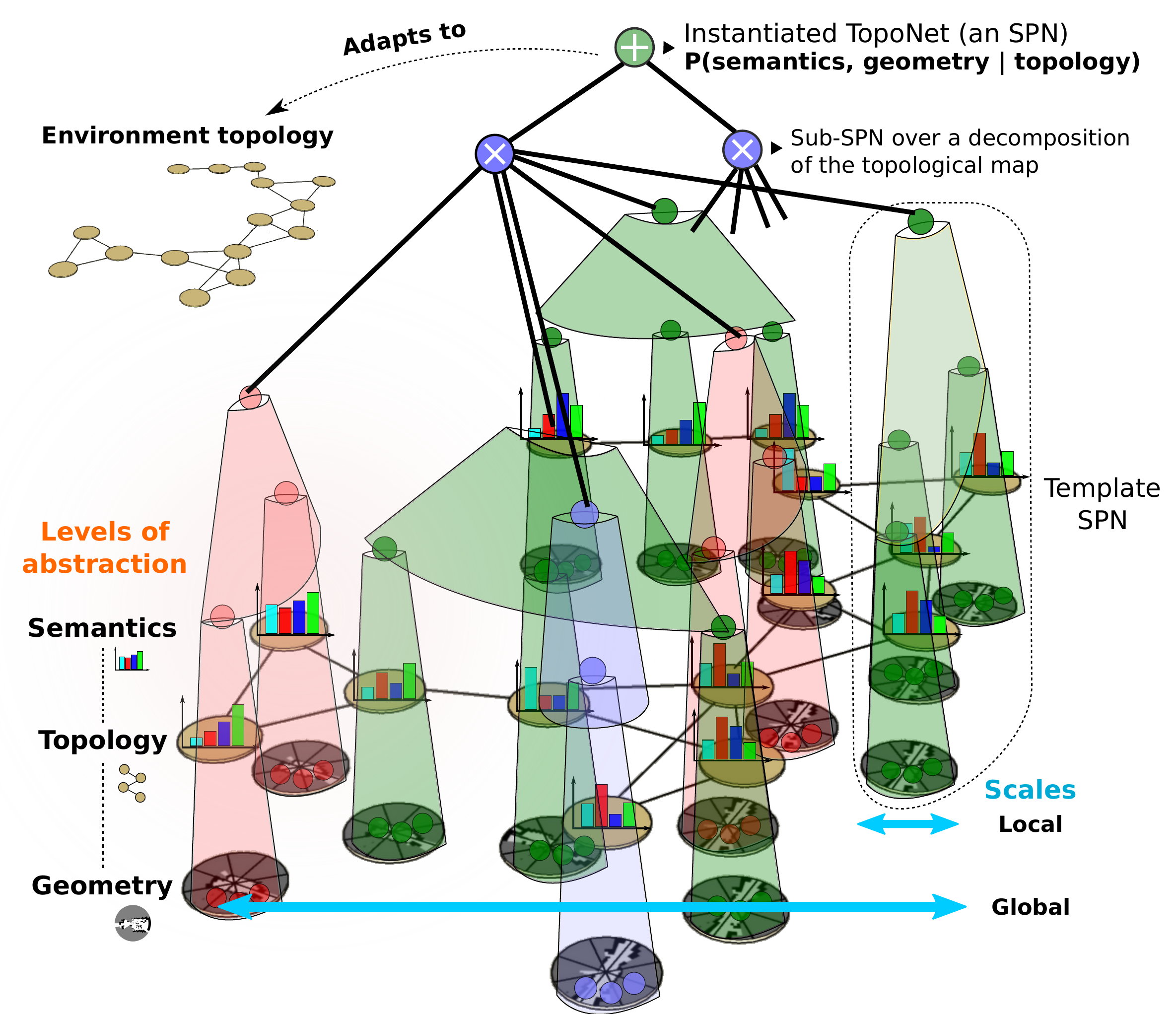}
  \caption{Illustration of a TopoNet instantiated over a semantic map, with
    structure adapted to the \kznotei{topology of the} environment. It
    incorporates spatial information across multiple levels of abstraction
    \kznotei{at} both local and global spatial scales, and forms a probability
    distribution over semantic attributes and \kznotei{geometric representations of
    places.}\\Video: \url{http://y2u.be/luv2XpaHeTU}.}
  \label{fig:pull}
\end{figure}

To overcome these shortcomings, in this work, we present TopoNets, end-to-end
deep networks for modeling semantic maps with dynamic structure adapted to
topology of large-scale environments. TopoNets leverage the \kznote{advantage}\kznotei{s}
of Sum-Product Networks (SPNs) and once \emph{instantiated}, form a unified
model that spans across abstractions and spatial scales, with guaranteed
tractable, exact inference (Fig.~\ref{fig:pull}).  In our experiments, we
evaluated TopoNets via the tasks of semantic place classification, inference of
semantics of unexplored places, and detection of novel environment
configurations. In each task, we compare \kznote{the end-to-end} TopoNets
with \kznote{a more traditional model assembled from two
  components: a Markov Random Field (MRF) \kznotei{representing} spatial relations and a deep
  model capturing place appearance.}
We show that TopoNets \kznotei{more effectively} disambiguate noisy \kznote{local}
predictions based on \kznote{real-world} observations, \kznotei{and} perform uncertain reasoning
about yet unexplored space.
\kznotei{Furthurmore, we demonstrate that} TopoNets exhibit generative
properties \kznote{useful for} novelty detection,
\kznotei{and achieve real-time performance while permitting exact probabilistic inference.}

\section{RELATED WORK}\label{sec:rel}

There have been numerous attempts to employ structured prediction to modeling
semantic maps with topological spatial relations. Mozos et
al.~\cite{mozos2007ras} used hidden Markov models (HMMs) to smooth sequences of
AdaBoost classifications of place observations into semantic
categories. Friedman et al. \cite{friedman2007ijcai} proposed Voronoi Random
Fields (VRFs) which are CRFs constructed according to a Voronoi graph extracted
from an occupancy grid map. VRFs utilize pairwise potentials to model dependency
between neighboring graph nodes and 4-variable potentials to model
junctions. Pronobis and Jensfelt \cite{pronobis2012icra} applied Markov Random
Fields to model pairwise dependencies between semantic categories of rooms
according to a topological map. The categorical variables were connected to
Bayesian Networks that reasoned about local environment features, forming a
chain graph. This approach relied on a door detector to segment the environment
into a topological graph with only one node per room. \kznotei{Overall, while}
probabilistic, \kznotei{these approaches} employ approximate inference, leading
to problems with convergence~\cite{friedman2007ijcai}. Moreover, additional
prior knowledge or hard commitments about the semantics of some places is
required in order to obtain a tractable model. In contrast, in this work, we
make no such commitments and rely on topological maps built by a real robot
while performing navigation and action execution. At the same time,
probabilistic inference with our model remains exact and real-time.

Recently several deep structured prediction methods have been proposed
\cite{mahmood2019structured}\cite{wu2016deep}\cite{chen2018deeplab}. Unfortunately,
most are designed for computer vision tasks and are not applicable to the
problem of modeling spatial relations in large-scale dynamic
environments. Notably, Mahmood et~al.~\cite{mahmood2019structured} proposed a
feature fusion method for conditional GAN which is conceptually similar to a
Conditional Random Field (CRF). The approach does not consider the joint
probability distribution of local observations and semantics as we do in this
work. Wu et~al.~\cite{wu2016deep} proposed a deep variant of MRFs based on
multiple recurrent neural networks for vision tasks. However, the method is
applicable only to problems with fixed number of variables, while our approach
handles graphs of arbitrary size and structure.

\kznote{TopoNets build upon our previous work, which introduced Sum-Product
  Networks (SPNs) to the domain of robotics.}
First, Pronobis
et~al.~\cite{pronobis2017iros} established the use of SPNs for local place
classifiction via a deep generative architecture that models robot-centric laser
range observations. Second, Zheng et~al.~\cite{zheng2018aaai} proposed a general
probabilistic approach to structured prediction, named GraphSPN, that
\kznote{extended SPNs to allow for the}
modeling of arbitrary, dynamic graphs.
\kznote{That work provided a \kznotei{new} theoretical framework, yet relied
  on synthetic local evidence.}
\kznotei{I}n this paper, \kznote{we} propose a unified, \kznote{end-to-end} architecture,
and experiment with \kznote{real-world} robot data collected in office settings
\kznote{to} demonstrate \kznote{its} practical value to the semantic mapping
problem.

\begin{figure}[ht!]
  \centering
  \includegraphics[width=0.7\linewidth]{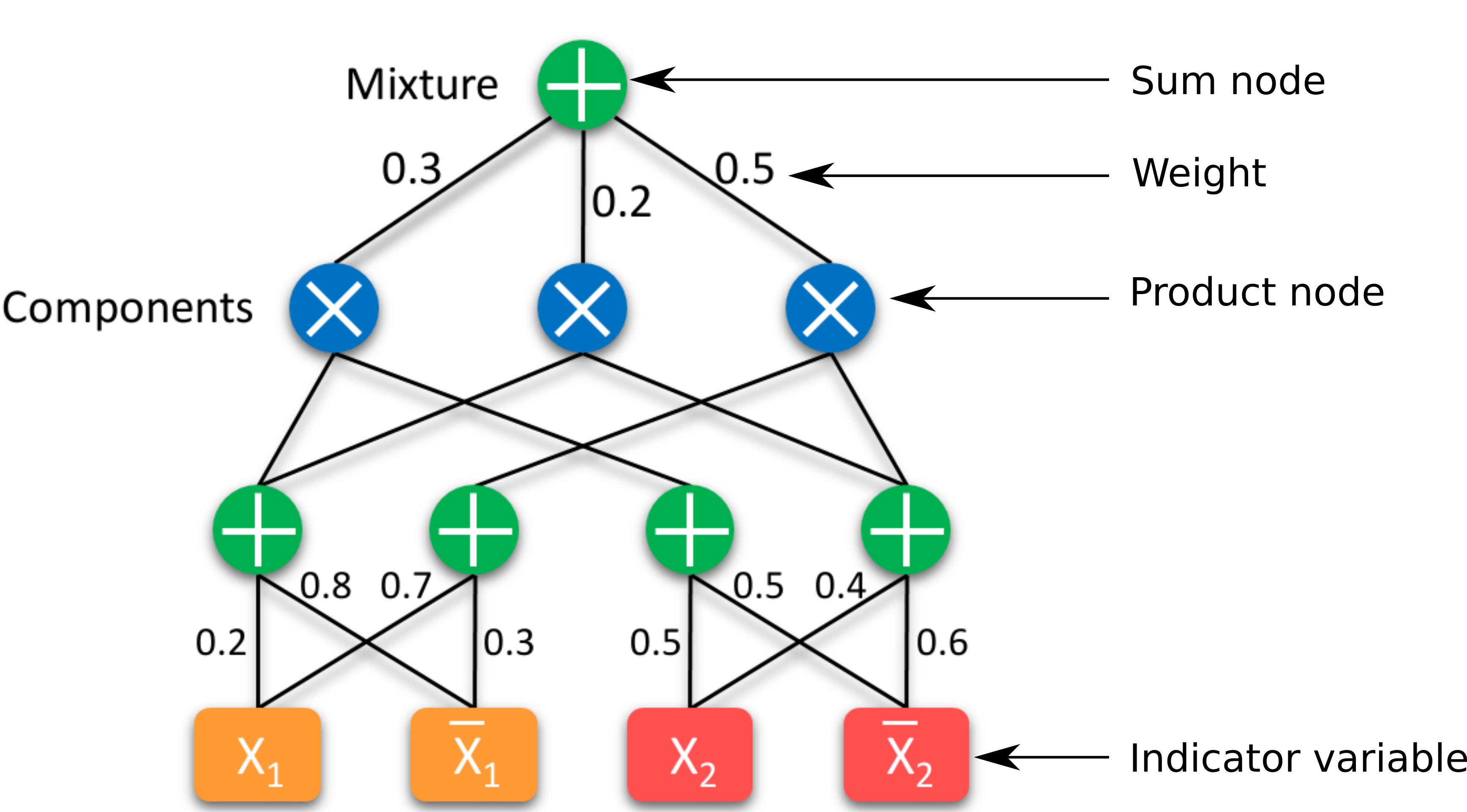}
  \caption{\label{fig:spn} A simple SPN for a naive Bayes mixture
  model $P(X_1,X_2)$, with three components over two binary variables. The
  bottom layer consists of indicators for different values of
  the variables $X_1$ and $X_2$. Weighted sum nodes,
  with weights attached to inputs, are marked with $+$, while product nodes are
  marked with~$\times$.}
\end{figure}

\section{PRELIMINARIES}

We begin by giving a brief introduction to Sum-Product Networks (SPNs), which
provide the fundamental theoretical framework for TopoNets. Then, we describe the
structure of the semantic maps for which TopoNets are built.

\subsection{Sum-Product Networks}\label{sec:spn}

SPNs are deep probabilistic models with solid theoretical
foundations~\cite{peharz2017pami}\cite{poon2011uai}\cite{gens2012nips} that have been shown
to provide state-of-the-art results in several
domains~\cite{pronobis2017iros}\cite{gens2012nips}\cite{peharz2014icassp}\cite{amer2015tpami}. One
of the primary limitations of traditional probabilistic graphical models is the
complexity of their partition function, often requiring complex approximate
inference in the presence of non-convex likelihood functions. In contrast, SPNs
represent probability distributions with partition functions that are guaranteed
to be tractable and involve a polynomial number of sum and product operations,
permitting exact inference. SPNs combine these advantages with benefits of deep
learning by acquiring hierarchical probabilistic models directly from
high-dimensional, noisy data.
While not all probability distributions can be encoded by polynomial-sized SPNs,
recent experiments in several domains show that the class of distributions
modeled by SPNs is sufficient for many real-world problems, including
speech~\cite{peharz2014icassp} and language
modeling~\cite{cheng2014interspeech}, human activity
recognition~\cite{amer2015tpami}, image classification~\cite{gens2012nips},
image completion~\cite{poon2011uai}, and robotics~\cite{pronobis2017iros}.

As shown in Fig.~\ref{fig:spn}, on a simple example of a naive Bayes
mixture model, an SPN is a generalized directed acyclic graph composed of
weighted sum and product operations. The sums can be seen as mixture models over
subsets of variables, with weights representing mixture priors.
Products can be viewed as features or mixture components. Not all possible
architectures consisting of sums and products result in valid probability
distributions and certain constraints (completeness and
decomposability~\cite{poon2011uai}\cite{peharz2015aistats}) must be followed to
guarantee validity.

SPNs model joint or conditional distributions and can be learned
generatively~\cite{poon2011uai} or discriminatively~\cite{gens2012nips} using
Expectation Maximization (EM) or gradient descent (GD). Additionally, several
algorithms were proposed for simultaneous learning of network parameters and
structure~\cite{poupart2017structure}\cite{gens2013icml}\cite{peharz2013mlkdd}. In this
work, we use a simple structure learning technique~\cite{pronobis2017iros} which
begins by initializing the SPN with \kznote{a} random dense structure that is later pruned.
The approach recursively generates network nodes based on multiple random
decompositions of the set of variables into multiple subsets until each subset
is a singleton. The resulting structure is a deep network consisting of products
combining the subsets in each decomposition and sums mixing different
decompositions at each level. SPNs can be defined for both continuous and
discrete variables, with evidence for categorical variables often specified in
terms of binary indicators.

Inference in SPNs is accomplished by an upwards pass which calculates the
probability of the evidence and a downwards pass which obtains gradients for
calculating marginals or MPE (Most Probable Explanation) state of the missing
evidence. The latter can be obtained by replacing sum operations with weighted
max operations (the resulting network is sometimes referred to as Max-Product
Network, MPN~\cite{gens2012nips}). For a detailed explanation of SPNs, we refer the reader to \cite{peharz2015aistats}\cite{gens2012nips}\cite{poon2011uai}.

\subsection{Semantic Maps}\label{sec:semantic_maps}
In order to represent dynamic spatial relations at the scale of a building, we
define semantic maps as \emph{growing} topological graphs of places associated with
observations of local geometry as well as semantic descriptions. Examples of the
topological and semantic information in such maps \kznote{acquired by a robot}
\kznote{(without local geometries)} are shown in Fig.~\ref{fig:instances}.
To obtain the representation of local place geometry, as the first step, we perform
spatio-temporal integration of the sensory input. We rely on laser-range data,
and use a particle-filter grid mapping~\cite{gmapping} to maintain a
robot-centric map of 5m radius around the robot. The goal of the local
representation is to model geometry of a single place. Thus, we constrain the
observation of a place to the information visible from the robot (structures
that can be raytraced from the robot's location). As a result, walls occlude the
view and the local map mostly contains information from a single room.

In our implementation, spatial relationships within each local place are modeled
from the perspective of a mobile robot acting at that place. Therefore, in the
next step, each local observation is transformed into a robot-centric polar
occupancy grid. \kznote{Examples
of such local place representations acquired by a robot can be seen in
Fig.~\ref{fig:classes}}.
The resulting observation contains
higher-resolution details closer to the robot and lower-resolution context
further away. This relates to how spatial information is used by a mobile robot
when planning and executing actions. It is in the vicinity of the robot that
higher accuracy of spatial information is required. In the future, we plan to use a similar
strategy when representing 3D and visual information, by extending the polar
representation to 3 dimensions.

The topological graph of a complete semantic map is built and updated
incrementally while the robot is exploring its
environment~\cite{pronobis2017icaps-planrob}. The primarily purpose of the graph
is to support the behavior of the robot. As a result, nodes in the graph
represent \emph{places} the robot can visit and the edges represent both navigability
and spatial relations. The places are associated with their local geometry
representations and latent variables representing semantics. Additional nodes in
the graph, called \emph{placeholders}, are created to represent exploration
frontiers. Those frontiers are added at neighboring, reachable, but unexplored
locations and connected to existing places. Then, once the robot performs an
exploration action, a placeholder is converted into a place, \kznote{to which a}
local \kznoteii{geometric} place representation is anchored.

\begin{figure*}[t!]
  \centering
  \captionsetup{width=\linewidth}
  \includegraphics[width=\linewidth]{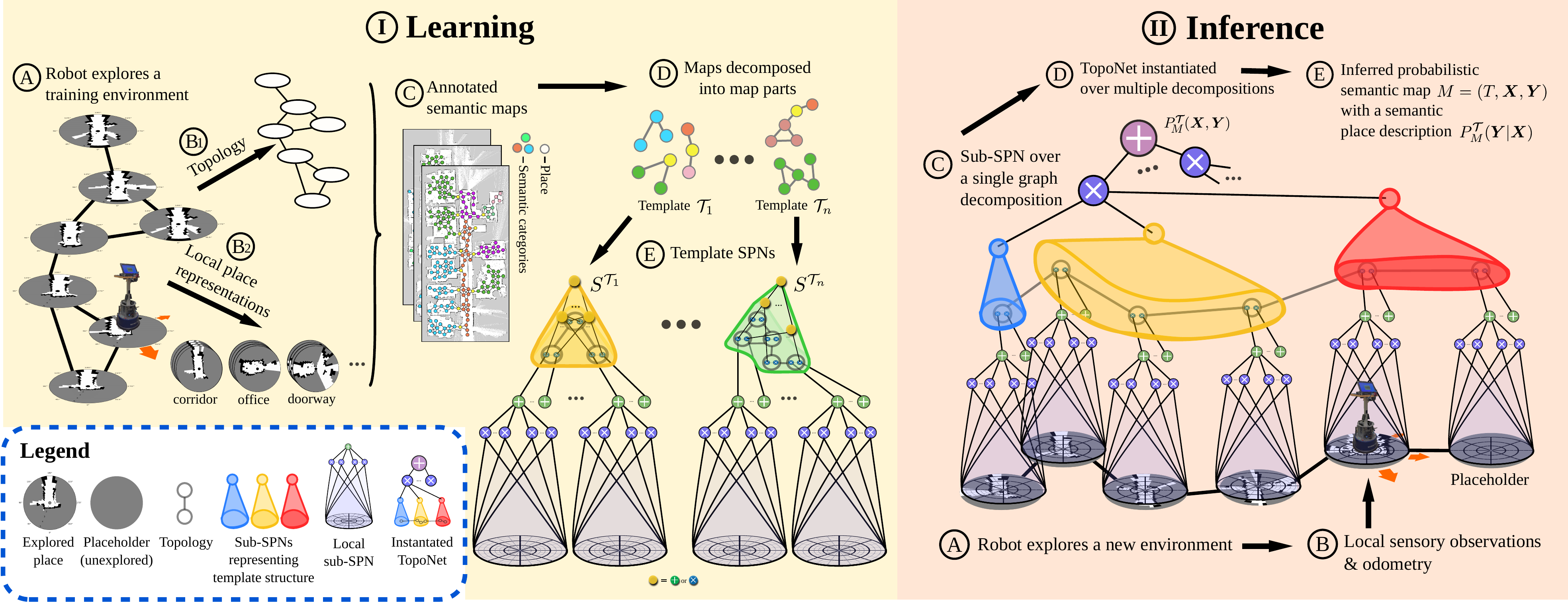}
  \caption{Learning and inference process with TopoNets.
    See step-by-step description in Sec.~\ref{sec:learn} and Sec.~\ref{sec:inf}.}
  \label{fig:newprocess}
\end{figure*}

\section{TOPONETS}\label{sec:toponet}

TopoNets are \kznote{deep} SPNs that adapt their structure according to the
topology and spatial relations in a semantic map. We begin \kznote{with a}
formal \kznote{definition} \kznotei{of} TopoNets, followed by
\kznote{a description of} the learning and inference procedure.

\subsection{Definition}\label{sec:topo_def}
Let us use $\bm{X}_i$ to denote local observations of place geometry,
and $\bm{Y}_i$ to denote semantic attributes that describe the
places.  We can specify a semantic map as $M=(T,\bm{X},\bm{Y})$, where
$T=(\bm{V},\bm{E})$ is a topological graph with vertices $\bm{V}$ and edges
$\bm{E}$, and $\bm{X}=\{\bm{X}_i: i \in \bm{V}\}$, $\bm{Y}=\{\bm{Y}_i: i \in
\bm{V}\}$.

A TopoNet is not specific to any particular semantic map, but rather a
template-based model that can be \emph{instantiated} for certain states of a semantic map to
perform inference tasks. To define TopoNets, we start by specifying a set
$\mathbfcal{T}=\{\mathcal{T}_1,\cdots,\mathcal{T}_n\}$ of \emph{sub-map templates}, which can be used to decompose a
semantic map. We define a \emph{sub-map template}
$\mathcal{T}=(\mathcal{V},\mathcal{E})$ as a graph with
$\mathbfcal{X}=\{\bm{X}_i: i\in\mathcal{V}\}$ and $\mathbfcal{Y}=\{\bm{Y}_i:
i\in\mathcal{V}\}$. Such template can be a recurring topological structure in a
given dataset of semantic maps. Following \cite{zheng2018aaai}, we define the
resulting decompositon as:
\begin{definition}\label{def:decomp}
  A decomposition of a semantic map $M=(T,\bm{X},\bm{Y})$ using
  {sub-map templates} ${\mathbfcal{T}}$ is a
  set of \emph{map parts} $M_k=(T_k, \bm{X}_k, \bm{Y}_k)$, with
  $T_k=(\bm{V}_k, \bm{E}_k)$, such that $T_k$ is isomorphic with any
  $\mathcal{T} \in {\mathbfcal{T}}$,
  $\bigcup_k T_k = T$, $\forall_{k,l} T_k \cap T_l = \varnothing$, and
  the variables $\bm{X}_k$ and $\bm{Y}_k$ correspond to vertices of $T_k$:
  $\bm{X}_k=\{\bm{X}_i: i \in \bm{V}_k\}$,
  $\bm{Y}_k=\{\bm{Y}_i: i \in \bm{V}_k\}$ .
\end{definition}

\kznote{\kznotei{With that}, we can define TopoNets as a template model
  consisting of a set of \emph{template SPNs}:}
\begin{definition}
  A \emph{template SPN} $\mathcal{S}^{\mathcal{T}}[\mathbfcal{X},\mathbfcal{Y}]$
  corresponding to \kznote{a} sub-map template $\mathcal{T}$ is an SPN that models the
  distribution $P^{\mathcal{T}}(\mathbfcal{X},\mathbfcal{Y})$.
\end{definition}
\begin{definition}
  A TopoNet $\mathbfcal{S}^{\mathbfcal{T}}$ is a set of \emph{template SPNs}
  such that
  $\mathbfcal{S}^{\mathbfcal{T}}=\{\mathcal{S}^{\mathcal{T}}[\mathbfcal{X},\mathbfcal{Y}]:\mathcal{T}\in\mathbfcal{T}\}$.
\end{definition}

\subsection{Learning (\kznote{Fig.~\ref{fig:newprocess}, I)}}\label{sec:learn}
As the robot explores a training environment (A), it can incrementally construct
a topological graph (B$_1$) using the approach described in
Sec.~\ref{sec:semantic_maps} and obtain local sensory observations at each place
(B$_2$). Eventually, the robot collects a dataset of annotated semantic maps
$\bm{M_{train}}$ (C). We \kznotei{specify} a set $\mathbfcal{T}$ of
\emph{sub-map templates}, each corresponding to a \emph{template SPN}. We use
$\mathbfcal{T}$ to decompose $\bm{M_{train}}$, which leads to a dataset of
\emph{map parts} \kznote{(D)}, defined in Def.~\ref{def:decomp}. Then, the
structure and parameters of \kznote{each} template SPN
\kznote{$\mathcal{S}^{\mathcal{T}}$ are} learned using the dataset of matching
map parts (E). \kznoteii{The specific procedure used to obtain template SPNs in}
\kznote{our experiments} is described in Sec.~\ref{sec:toponet_realize}.


\begin{figure}[t]
  \centering
  \captionsetup{width=\linewidth}
  \includegraphics[width=\linewidth]{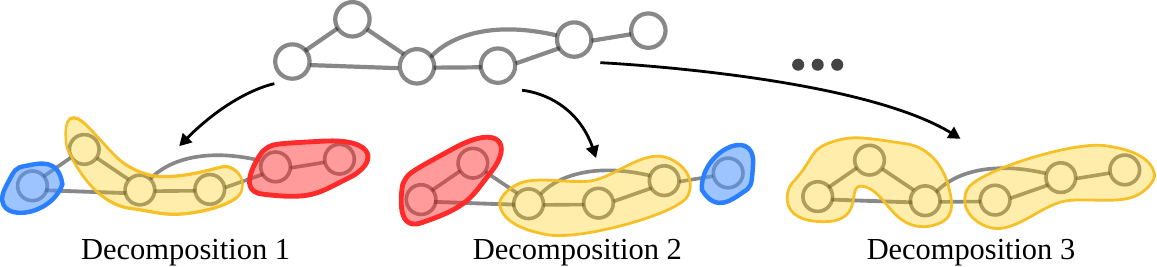}
  \caption{A topological graph can be decomposed in \kznotei{multiple} ways
    using the same \kznotei{set of} sub-map
    templates.}\label{fig:graph_decompose}
\end{figure}

\subsection{Inference (\kznote{Fig.~\ref{fig:newprocess}, II)}}\label{sec:inf}

When the robot explores a new \kznotei{test} environment (A), it constructs a
topological graph with each place corresponding to a local geometry
representation (B). Next, a trained TopoNet $\mathbfcal{S}^{\mathbfcal{T}}$ is
adapted to the topological structure of the underlying semantic map
\kznotei{$\bm{M}_{test}(T,\bm{X},\bm{Y})$} and latent place semantics is
inferred. This adaptation process is the \emph{instantiation} of a TopoNet
\kznotei{\kznoteii{performed} as follows}.  \kznote{First, the
  \kznoteii{semantic map} is decomposed into \emph{map parts} using sub-map
  templates $\mathbfcal{T}$. For each map part $M_k=(T_k, \bm{X}_k, \bm{Y}_k)$,
  a template SPN $\mathcal{S}^{\mathcal{T}}\in\mathbfcal{S}^{\mathbfcal{T}}$ is
  selected such that $T_k$ is isomorphic with $\mathcal{T}$. The structure and
  weights of $\mathcal{S}^{\mathcal{T}}$ are \kznotei{instantiated as an SPN}
  $\mathcal{S}^{\mathcal{T}}[\bm{X}_k,\bm{Y}_k]$ that models the distribution
  $P^{\mathcal{T}}(\bm{X}_k,\bm{Y}_k)$.  The instantiated template SPNs for all
  map parts are combined with a product node, which \kznotei{forms} a sub-SPN
  over a single, \kznotei{complete} graph decomposition (C).  Using the same
  $\mathbfcal{T}$, a topological graph \kznotei{is} decomposed in
  \kznotei{multiple} different ways (Fig.~\ref{fig:graph_decompose}).}
After $N$ \emph{different} decomposition attempts, \kznote{the} $N$ product
nodes become the children of the root sum node of the \kznotei{final} network (D). This
forms a distribution \kznoteip{$P^{\mathbfcal{T}}_{\bm{M}_{test}}(\bm{X},\bm{Y})$}, which can be
seen as a mixture model over the different decompositions.

Once instantiated, TopoNets can perform different types of
\kznote{probabilisitic} inferences\kznotei{:}
\subsubsection{Semantic place classification}\label{sec:inf_semantics}
\kznotei{For all places explored by the robot, \kznoteip{where} local observations $\bm{X}$
  are available, we \kznoteii{can} task TopoNets with inferring latent semantics:
\begin{align}
\hat{\bm{y}}_{explored} = \argmax_{\bm{y}_{explored}}P(\bm{y}_{explored}|\bm{x}_{explored})
\end{align}}

\subsubsection{Inferring semantics of unexplored space}\label{sec:inf_placeholder}
\kznotei{We \kznoteii{can} increase the complexity of the task and infer semantic descriptions
  of both explored places and nearby unexplored placeholders, for which local
  evidence is not available: 
  \begin{align}
    \begin{split}
      &\hat{\bm{y}}_{explored},\hat{\bm{y}}_{unexplored}\\
      &\qquad= \argmax_{\substack{\bm{y}_{explored};\\\bm{y}_{unexplored}}}P(\bm{y}_{explored},\bm{y}_{unexplored}|\bm{x}_{explored})
    \end{split}
  \end{align}}

\subsubsection{Novelty detection}\label{sec:inf_novelty}
\kznotei{This inference task evaluates the generative properties of TopoNets. For a
certain set of local observations $\bm{x}_{explored}$, we can evaluate the likelihood
$P(\bm{x}_{explored})$ and use it as a measure of novelty. The likelihood \kznoteii{can be} thresholded
to determine \kznoteii{whether} the complete environment is within the distribution of
environments known during training: 
\begin{align}
\sum_{\bm{y}_{explored}} P(\bm{y}_{explored},\bm{x}_{explored}) > threshold
\end{align}}

\begin{figure*}[t]
  \centering
  \captionsetup{width=\linewidth}
  \includegraphics[width=\linewidth]{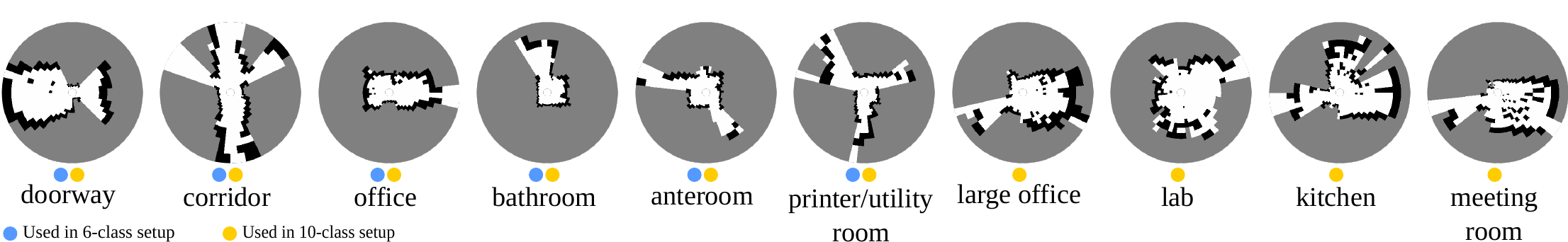}
  \caption{Examples of local geometry observations for places of different groundtruth categories. }\label{fig:classes}
\end{figure*}

\section{EXPERIMENTAL SETUP}\label{sec:setup}

\subsection{Dataset}

Our experiments were performed for semantic maps built from laser-range and
odometry data from the COLD-Stockholm
dataset\footnote{\url{http://coldb.org}}. The dataset contains 32 data
sequences captured using a mobile robot navigating 
\kznote{four} floors (floors 4-7) of an office building. On each floor,
the robot \kznoteii{explored} 
rooms of different semantic categories.  We
experimented with two place category setups, \kznote{with 6 and 10 classes,}
shown in Fig.~\ref{fig:classes}. \kznote{Each} class \kznoteii{appeared} 
on at least two of~the four floors.
We \kznoteii{used} \kznote{the two setups} to \kznote{illustrate}
how TopoNets behave in settings of varying difficulty.
To ensure variability between training and test sets, we split the
\kznote{dataset} 
four times, each time training TopoNets on data from three floors and leaving
one floor out for testing. Note that unlike in \cite{zheng2018aaai}, where
experiments were conducted with synthetic observations, we \kznoteii{experimented}
with real data and \kznote{more challenging} configurations than the
4-class setup in~\cite{pronobis2017iros}. 

\subsection{\kznote{Realization of TopoNets}}\label{sec:toponet_realize}

\begin{figure}[tb]
  \centering
  \includegraphics[scale=0.65]{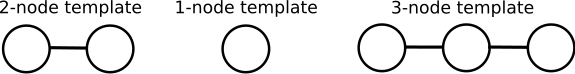}
  \caption{Sub-map templates used in our experiments.}\label{fig:templates}
\end{figure}


\kznotei{The TopoNets framework can be adapted to the complexity of the
  topological maps and the type of sensory input available to the robot. In our
  experiments, we built TopoNets using a set of three simple sub-map templates
  shown in Fig.~\ref{fig:templates}. For each sub-map template, the structure
  and weights of the corresponding template SPN were obtained using the same
  protocol \kznoteii{described} \kznoteip{as follows.}}

\kznotei{The structure of each template SPN was partially designed based on
  domain knowledge and partially learned according to the algorithm described in
  Sec.~\ref{sec:spn}. At the bottom of the network, the geometry of each place
  in the sub-map template \kznoteii{was} modeled independently, using sub-SPNs
  \kznoteii{resembling} the place classification model proposed in
  \cite{pronobis2017iros}. For each place, the local sensory input integrated
  into a polar occupancy grid was captured using a set of indicator variables.
  The resolution of the local place geometries (Fig.~\ref{fig:classes})
  \kznoteii{was} 56 angular cells by 21 radial cells, resulting in 1176 random
  variables per place. \kznoteii{Next, we split} the polar occupancy grid
  equally into eight $45$-degree views. For each view, we \kznoteii{learned} an
  independent sub-SPN. This \kznoteii{allowed} us to use networks of different
  complexity for representing low-level features and high-level structure of a
  place. On top of the sub-SPNs representing the views, we \kznoteii{learned} a
  sub-SPN representing a complete place geometry. That process \kznoteii{was}
  repeated for each semantic place class, resulting in either 6 or 10 sub-SPNs
  for each place in the sub-map template. The upper layers of the template SPN
  \kznoteii{combined} all the sub-SPNs into a single template distribution. The
  structure of those layers \kznoteii{was} \kznoteip{initialized to} best
  represent the characteristics of a specific sub-map template \kznoteip{and
    learned as described in Sec.~\ref{sec:spn}}.}




The parameters \kznotei{of the model} were learned using Gradient Descent via
two different losses for different layers of a template SPN. For \kznotei{the
  bottom} sub-SPNs \kznotei{representing features of specific semantic classes,}
we employed a cross-entropy discriminative loss in order to
maximize classification performance. \kznotei{In contrast,} the top \kznotei{layers}
\kznoteip{were trained with} Maximum-Likelihood generative loss in order to retain good
generative abilities and estimation of likelihoods for complete semantic
maps. This provided a good trade-off between the different abilities of the
probabilistic representation.

\begin{table*}[t]
  \centering
\resizebox{\textwidth}{!}{%
\begin{tabular}{rcrrrrrr}
\multicolumn{8}{c}{A. Semantic Place Classification}                                                                                                                                                                               \\ \hline\hline    
\multicolumn{1}{c}{\multirow{2}{*}{Data Split}}           & \multirow{2}{*}{\#classes} & \multicolumn{2}{c}{Local SPNs}                  & \multicolumn{2}{c}{Local SPNs + MRF}            & \multicolumn{2}{c}{TopoNet}                       \\
                                                         &                            & \multicolumn{1}{c}{avg.} & \multicolumn{1}{c}{std.} & \multicolumn{1}{c}{avg.} & \multicolumn{1}{c}{std.} & \multicolumn{1}{c}{avg.} & \multicolumn{1}{c}{std.} \\ \hline\hline
\multicolumn{1}{c}{\multirow{2}{*}{456-7}}                & 6                          & 95.22\%                  & 1.70\%                   & 96.35\%                  & 2.68\%                   & \textbf{97.50\%}         & 1.11\%                   \\
                                                         & 10                         & 73.48\%                  & 1.92\%                   & 68.03\%                  & 4.55\%                   & \textbf{74.69\%}         & 2.73\%                   \\ \hline
\multicolumn{1}{c}{\multirow{2}{*}{457-6}}                & 6                          & 96.75\%                  & 1.98\%                   & 93.87\%                  & 2.24\%                   & \textbf{97.39\%}         & 1.25\%                   \\
                                                         & 10                         & 81.49\%                  & 1.93\%                   & 81.63\%                  & 6.90\%                   & \textbf{82.55\%}         & 1.37\%                   \\ \hline
\multicolumn{1}{c}{\multirow{2}{*}{467-5}}                & 6                          & 92.70\%                  & 1.52\%                   & \textbf{95.66\%}         & 3.14\%                   & 94.46\%                  & 1.62\%                   \\
                                                         & 10                         & 73.41\%                  & 2.06\%                   & 66.63\%                  & 5.38\%                   & \textbf{74.58\%}         & 2.48\%                   \\ \hline
\multicolumn{1}{c}{\multirow{2}{*}{567-4}}                & 6                          & \textbf{99.16\%}         & 0.94\%                   & 96.00\%                  & 3.21\%                   & 98.30\%                  & 1.64\%                   \\
                                                         & 10                         & 87.88\%                  & 2.83\%                   & 84.31\%                  & 1.56\%                   & \textbf{88.73\%}         & 2.35\%                   \\ \hline\hline
\multicolumn{1}{c}{\multirow{2}{*}{Overall}}              & 6                          & 95.96\%                  & 2.83\%                   & 95.47\%                  & 3.00\%                   & \textbf{96.91\%}         & 2.04\%                   \\
                                                         & 10                         & 79.06\%                  & 6.45\%                   & 75.15\%                  & 9.34\%                   & \textbf{80.14\%}         & 6.35\%                   \\ \hline\hline
\end{tabular}
\qquad
\begin{tabular}{rcrrrr}
\multicolumn{6}{c}{B. Inferring Semantics of Unexplored \kznoteii{Space}}                                                              \\ \hline\hline
\multicolumn{1}{c}{\multirow{2}{*}{Data Split}}   & \multirow{2}{*}{\#classes} & \multicolumn{2}{c}{Local SPNs + MRF} & \multicolumn{2}{c}{TopoNet} \\
                                                  &           & avg.            & std.          & avg.           & std.         \\ \hline\hline
\multicolumn{1}{c}{\multirow{2}{*}{456-7}}        & 6         & 94.07\%         & 5.65\%        & \textbf{99.46\%}        & 1.44\%       \\      
                                                  & 10        & 57.94\%         & 3.33\%        & \textbf{70.24\%}        & 10.18\%      \\ \hline
\multicolumn{1}{c}{\multirow{2}{*}{457-6}}        & 6         & 79.77\%         & 6.49\%        & \textbf{83.29\%}        & 4.26\%       \\      
                                                  & 10        & \textbf{61.62\%}& 9.48\%        & 61.30\%        & 4.83\%       \\ \hline
\multicolumn{1}{c}{\multirow{2}{*}{467-5}}        & 6         & 94.72\%         & 3.48\%        & \textbf{96.35\%}        & 3.62\%       \\      
                                                  & 10        & 50.50\%         & 5.48\%        & \textbf{54.48\%}        & 3.38\%       \\ \hline
\multicolumn{1}{c}{\multirow{2}{*}{567-4}}        & 6         & 96.09\%         & 3.50\%        & \textbf{98.75\%}        & 3.31\%       \\      
                                                  & 10        & 71.72\%         & 4.78\%        & \textbf{71.94\%}        & 8.45\%       \\ \hline\hline
\multicolumn{1}{c}{\multirow{2}{*}{Overall}}      & 6         & 91.16\%         & 8.27\%       & \textbf{94.46\%}        & 7.35\%       \\
                                                  & 10        & 60.45\%         & 9.84\%        & \textbf{64.49\%}        & 10.11\%      \\ \hline\hline
\end{tabular}
}
\caption{Results of the experiments with semantic place classification and inference of semantics of unexplored space.}
 \label{tab:inf_results}
\end{table*}

\subsection{Baseline}
\kznote{As discussed in Section \ref{sec:rel}, existing deep approaches to
  structured prediction and semantic mapping \kznoteii{could not} be directly applied to the
  task formulated in this paper. Therefore, as a baseline, we \kznoteii{used} a
  more traditional model composed of two sub-models. First, similarly to the
  semantic mapping techniques in
  \cite{pronobis2012icra}\cite{posner2009generative}, we \kznoteii{used} a
  pairwise Markov Random Field (MRF) to capture dynamic spatial relations in
  topological graphs. However, since the complexity of the sensory observations
  requires a different perceptual model, we \kznoteii{combined} the MRFs with a
  deep representation capturing the geometry of local places. To this end, we
  \kznoteii{emploed} local SPN models, structured identically to the bottom
  layers of \kznotei{our} TopoNets, and trained \kznoteii{them} discriminatively
  to infer semantic place categories of independent places based only on local
  evidence. The resulting model \kznoteii{used} evidence from local SPNs as
  unary potentials $\phi_i(Y_i=c) = P(\bm{X}_i|Y_i=c)$ in the MRF. The pairwise
  potentials \kznoteii{were} obtained as in \cite{zheng2018aaai} by computing
  co-occurrence statistics of semantic classes of neighboring places in the
  training topological graphs.}


\subsection{Software and Performance}\label{sec:software}
The experiments \kznote{with} TopoNets were conducted using
\emph{LibSPN}~\cite{pronobis2017icml-padl}\footnote{\url{https://libspn.org}.},
a library for learning and inference with SPNs and TensorFlow, \kznote{as well
  as an implementation of}
GraphSPNs~\cite{zheng2018aaai}\footnote{\url{https://github.com/zkytony/graphspn}}.
For MRF experiments, we \kznoteii{used} an implementation of \kznote{Loopy Belief
  Propagation} \kznote{provided by} the \emph{libDAI} library
\cite{mooij2010jmlr}. We \kznote{compared} the inference time for TopoNets
\kznote{and the baseline} on semantic maps built for 10 classes.
\kznotei{TopoNets were} \kznote{built for} 40 decompositions of the semantic
maps. \kznote{For maps \kznotei{containing} 105 and 155 nodes},
TopoNets \kznoteii{evaluated} \kznoteip{$P^{\mathcal{T}}_{\bm{M}_{test}}(\bm{X},\bm{Y})$}
in 0.36s and 0.49s respectively, on a desktop computer with one GeForce GTX 1080
Ti GPU. \kznote{In comparison, inferences with MRF often required more than 45s
  due to poor convergence and hard-stopping.}
Note that \kznotei{these run times} \kznoteii{correspond to} inference
\kznotei{over} the entire semantic map. \kznotei{In practice, inference
  \kznoteii{with TopoNets} can be restricted to only \kznoteii{those} parts of the network
  \kznoteii{affected} by new evidence, drastically reducing the amount of required
  computations.}


\begin{figure*}[!bt]
  \centering
  \captionsetup{width=\linewidth}
  \includegraphics[width=\linewidth]{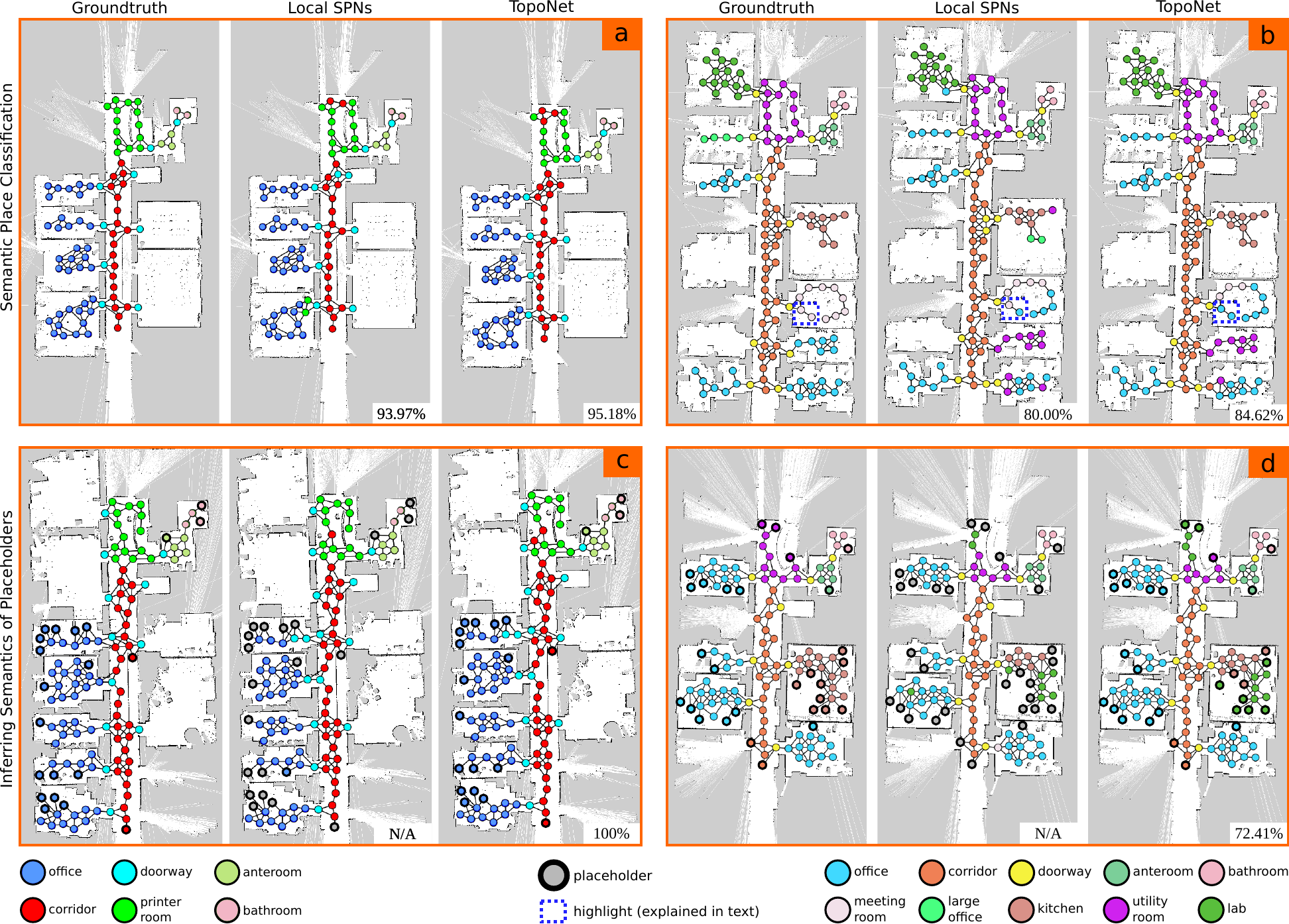}
  \caption{Visualization of TopoNet inference \kznoteii{results for four} sequences on
    different floors.
    The occupancy grid \kznoteii{maps} and \kznoteii{the} topological graphs
    \kznoteii{were} collected as the robot \kznoteii{navigated} the
    \kznoteii{environment}. The top row shows results for semantic
    \kznoteii{place} classification, \kznoteii{while} the bottom row
    \kznoteii{shows results} for inference of \kznoteii{unexplored space}. The
    accuracy for the corresponding tasks is shown \kznoteii{in the} bottom-right
    \kznoteii{corners}. Colors indicate the \kznoteii{groundtruth or the} most
    likely \kznoteii{inferred class}.}\label{fig:instances}
\end{figure*}

\begin{figure}[!tb]
  \centering
  \captionsetup{width=\linewidth}
  \includegraphics[width=\linewidth]{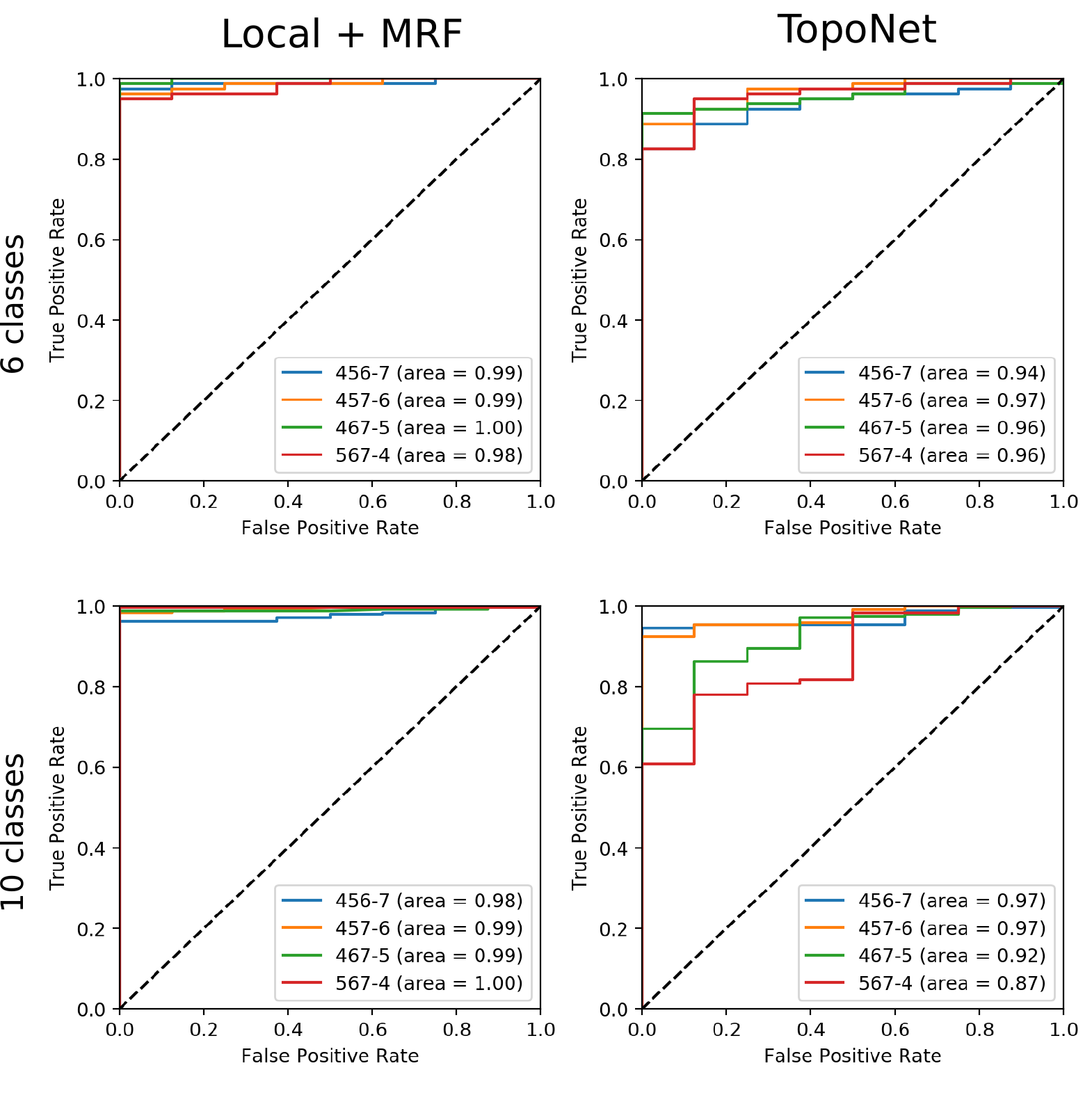}
  \caption{ROC plots for the task of novelty detection.}\label{fig:rocs}
\end{figure}

\section{RESULTS AND DISCUSSIONS}\label{sec:result}

\kznoteii{Below, we describe and discuss the results of three experiments corresponding to each of the
inference tasks specified in Sec.~\ref{sec:inf}}.


\subsubsection{Semantic Place Classification}\label{sec:res_classification}
First, we tasked TopoNets and the baseline employing MRFs and local SPNs with
inferring the semantics of explored places given local sensory observations
(Sec.~\ref{sec:inf_semantics}).  For this experiment, we used an additional
baseline consisting of independent local SPNs, inferring semantic
descriptions of independent places, without relying on topological spatial
relations. The accuracy for each experiment was calculated as the percentage
of all places in a test map for which the most likely inferred semantic class
matched the groundtruth.

As shown in TABLE~\ref{tab:inf_results}-A, local SPNs obtained overall accuracy
(over all test maps and data splits) of $95.96\%(\pm 2.83)$ for the 6-class
setup and $79.06\%(\pm 6.45)$ for the 10-class setup. By incorporating the
topological spatial relations, TopoNets improved that result to
$96.91\%(\pm 2.04)$ and $80.14\%(\pm 6.35)$, respectively. At the same time, the
solution employing MRFs for capturing spatial relations resulted in lower
performance in both cases: $95.47\%(\pm 3.00)$ and $75.15\%(\pm 9.34)$. This
trend was confirmed for most data splits, with TopoNets outperforming each
baseline in 7 out of 8 cases, and MRFs lowering the performance compared to
local SPNs in 5 out of 8 cases. In-depth analysis of the inference results
revealed that the likelihoods of the latent semantic categories calculated
independently for local places can be noisy~\cite{zheng2018aaai}. This
significantly impacted the performance of MRFs, while TopoNets remained largely
unaffected by noisy local evidence.

As shown in the visualizations of place classification results in
Fig.~\ref{fig:instances}a-b, local SPNs provided a strong baseline (particularly in
the 6-class setup). However, relying only on local evidence can often lead to
perceptual aliasing and misclassification of a cluster of nearby places. That
effect was more pronounced in the 10-class setup. In certain cases, the
misleading local evidence overpowered the global structural information. This
typically occurred in situations where the incorrect, alternative explanation for
the local evidence agreed with the global environment structure captured in the
training data (e.g.~a~meeting room misclassified as an office as shown in the
highlighted area in Fig.~\ref{fig:instances}b). In such a case, TopoNets could
spread the misclassifications to nearby places within the same room. However, in
most cases, TopoNets were able to exploit topological spatial relations to
correct misclassifications, resulting in improvement in the overall accuracy.

\subsubsection{Inferring Semantics of Unexplored Space}
Next, we tasked TopoNets and the baseline with inference of semantic
descriptions of unexplored placeholders lacking local evidence, based solely on
observations attached to adjacent explored places
(Sec.~\ref{sec:inf_placeholder}). Importantly, the semantics of the explored
places was not provided and remained latent in this experiment. The accuracy was
defined as the percentage of all placeholders in a test map for which the most
likely semantic class matched the groundtruth. We report the results in
TABLE~\ref{tab:inf_results}-B.

In this experiment, TopoNets outperformed the baseline even more significantly.
TopoNets correctly inferred the semantics of $94.46\%(\pm 7.35)$ and
$64.49\%(\pm 10.11)$ placeholders in the 6- and 10-class settings, respectively,
compared to $91.16\%(\pm 8.27)$ and $60.45\%(\pm 9.84)$ for the baseline. In~
fact, MRFs coupled with local deep models outperformed TopoNets in only 2 out of
32 sequences in the 6-class setting, and 8 out of 32 sequences in the 10-class
setting. As visualized in Fig.~\ref{fig:instances}c-d, TopoNets could
successfully exploit the knowledge about global environment structure to
distribute evidence
to unexplored space.

We observed that increasing the number of decompositions, had a positive
influence on the performance of TopoNets on this task ($92.43\%(\pm 8.08)$ for
12 decompositions and $94.46\%$ for 40 decompositions in the 6-class setting).
Placeholders are exploration frontiers that reside at the outer perimeter of the
semantic map, and a larger number of decompositions increases the chance of
placeholders being covered by complex sub-map templates.

These results together with the computational efficiency of TopoNets
(Sec.~\ref{sec:software}) illustrate their practical benefits for
spatial understanding in the open world.

\subsubsection{Novelty Detection}
Finally, we exploited the generative properties of the models to determine
whether whole environments match the distribution obtained during training or
can be considered novel. This property is particularly important for robots
operating in open, unknown environments.

For this experiment, we required evidence gathered in novel environments, which
were incongruent with the training data. To obtain novel semantic maps, we
randomly selected pairs of groundtruth classes, and swapped the local evidence
belonging to the two classes for all places in the test maps. For example, the
evidence for rooms labeled as an office was swapped with the evidence for
bathrooms, effectively creating new environments with different local
geometries, where offices now appeared to be bathrooms, and vice versa.  We
randomized 10 different novel maps for the 6-class setup and 30 for the 10-class
setup. At the same time, the original test set provided maps that, while
previously unseen, were considered to be within the distribution of the training
environments. Note that some of the random swaps generated environment
configurations that were similar to those in the training data (e.g. a kitchen
swapped with a meeting room), resulting in a difficult detection problem.

The novelty detection results are shown as ROC curves in
Fig.~\ref{fig:rocs}. Both approaches performed well on this task, with MRFs
outperforming TopoNets (average AUC of 0.99 for MRFs and 0.96 for TopoNets in
the 6-class case), a similar result to the one reported in~\cite{zheng2018aaai},
despite differences in the setup. From the plots, we see that when taking into
account all data splits and test maps, TopoNets were able to correctly detect
$89\%$ and $86.25\%$ of novel maps, in 6 and 10-class setups, respectively,
while missclassifying as false positives only $9.38\%$ of test maps for the
6-class case and $15.63\%$ for the 10-class case. Such level of performance can
be sufficient for many real-world applications, where novelty detection can be
used to avoid errors and trigger additional learning.

\section{CONCLUSIONS}
\kznotei{This paper presents TopoNets}, end-to-end deep networks for modeling
semantic maps with structure \kznoteii{adapting} to \kznotei{dynamic}
environment topology. Through experiments with \kznotei{real-world} robot
sensory observations, we comprehensively evaluated and analyzed the inference
behavior and generative properties of TopoNets. \kznoteii{We demonstrated that
  TopoNets are an efficient and practical approach to spatial
  understanding. Furthermore, their properties make them ideal for supporting
  behavior planning and execution in robots operating in large, open, unknown
  environments.}  \kznotei{It is our hope that \kznoteii{showcasing} the
  benefits of SPN-based \kznoteii{deep models} will provide a new direction for
  research towards novel probabilistic inference techniques in
  robotics.}

\bibliographystyle{IEEEtran}
\bibliography{IEEEabrv,root}

\end{document}